\documentclass[lettersize,journal]{IEEEtran}
\usepackage{amsmath,amsfonts}
\usepackage{algorithmic}
\usepackage{algorithm}
\usepackage{array}
\usepackage[caption=false,font=normalsize,labelfont=sf,textfont=sf]{subfig}
\usepackage{textcomp}
\usepackage{stfloats}
\usepackage{url}
\usepackage{verbatim}
\usepackage{graphicx}
\usepackage{cite}
\usepackage{multirow}
\usepackage{color}
\usepackage{amssymb}
\usepackage{booktabs}
\usepackage{newfloat}
\usepackage{listings}
\usepackage{flushend}
\usepackage{bibentry}
\begin{document}

\title{The Language of Touch: Translating Vibrations into Text with Dual-Branch Learning}

\author{Jin Chen, Yifeng Lin, Chao Zeng, Si Wu, and Tiesong Zhao, \IEEEmembership{Senior Member, IEEE}
	
	\IEEEcompsocitemizethanks{
    	This work is supported by the National Science Foundation of China (Grant No. 62571131) \textit{(Corresponding author: Tiesong Zhao.)}
 
        \IEEEcompsocthanksitem J. Chen and Y. Lin are with the Fujian Key Lab for Intelligent Processing and Wireless Transmission of Media Information, Fuzhou University, Fuzhou 350108, China(e-mails: \{231110036, 241120007\}@fzu.edu.cn).
        \IEEEcompsocthanksitem T. Zhao is with the Fujian Key Lab for Intelligent Processing and Wireless Transmission of Media Information, Fuzhou University, Fuzhou 350108, China and also with the Fujian Science and Technology Innovation Laboratory for Optoelectronic Information of China, Fuzhou 350108, China (e-mails: \{t.zhao\}@fzu.edu.cn).        
        \IEEEcompsocthanksitem C. Zeng is with the School of Artificial Intelligence, Hubei University, Wuhan, China, and the Key Laboratory of Intelligent Sensing System and Security, Hubei University and Ministry of Education, China. (email: chao.zeng@hubu.edu.cn)
        \IEEEcompsocthanksitem S. Wu is with the School of Computer Science and Engineering, South China University of Technology. (email: cswusi@scut.edu.cn)

}}

\markboth{Journal of \LaTeX\ Class Files,~Vol.~14, No.~8, August~2021}%
{Shell \MakeLowercase{\textit{et al.}}: A Sample Article Using IEEEtran.cls for IEEE Journals}


\maketitle

\begin{abstract}
The standardization of vibrotactile data by IEEE P1918.1 workgroup has greatly advanced its applications in virtual reality, human-computer interaction and embodied artificial intelligence. Despite these efforts, the semantic interpretation and understanding of vibrotactile signals remain an unresolved challenge. In this paper, we make the first attempt to address vibrotactile captioning, {\it i.e.}, generating natural language descriptions from vibrotactile signals. We propose Vibrotactile Periodic-Aperiodic Captioning (ViPAC), a method designed to handle the intrinsic properties of vibrotactile data, including hybrid periodic-aperiodic structures and the lack of spatial semantics. Specifically, ViPAC employs a dual-branch strategy to disentangle periodic and aperiodic components, combined with a dynamic fusion mechanism that adaptively integrates signal features. It also introduces an orthogonality constraint and weighting regularization to ensure feature complementarity and fusion consistency. Additionally, we construct LMT108-CAP, the first vibrotactile-text paired dataset, using GPT-4o to generate five constrained captions per surface image from the popular LMT-108 dataset. Experiments show that ViPAC significantly outperforms the baseline methods adapted from audio and image captioning, achieving superior lexical fidelity and semantic alignment. 
\end{abstract}


\section{Introduction}
Haptics plays a critical role in enhancing multimodal interaction by complementing visual and auditory modalities. It has been widely applied in industrial automation, telemedicine, hazardous environment exploration, and e-commerce. Typically, haptic signals can be classified into kinesthetic and vibrotactile components \cite{8816110}: the former involves motion, force, and torque, while the latter conveys surface-related properties such as friction, hardness, temperature, and roughness. Until now, vibrotactile signals can be captured by different types of sensors, including visuotactile sensors \cite{10563188} and triaxial accelerometers \cite{9807441}. Although visuotactile sensors such as GelSight have attracted considerable attention from the computer vision community \cite{8953737}, the IEEE P1918.1 Working Group standardized the representation of vibrotactile signals as multiple 1D time series in 2019 \cite{8605315}. This step has significantly enhanced the perception, transmission, reproduction, and cross-platform application of vibrotactile signals \cite{10452829}. It has also promoted interoperability across tactile devices and enabled system-level integration in virtual reality, human–computer interaction, and embodied artificial intelligence applications \cite{8605315}.

\begin{figure}[t!]
\centering
\includegraphics[width=0.45\textwidth]{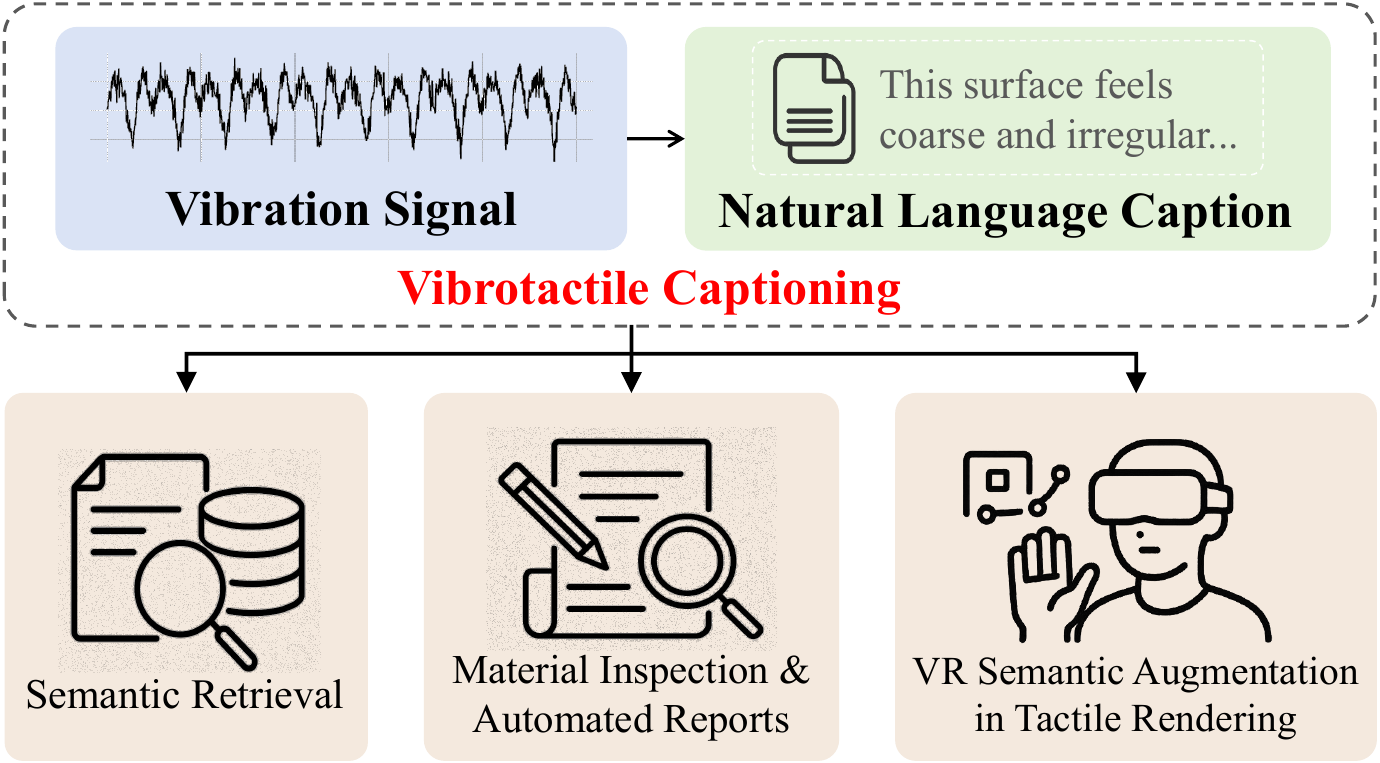} 

\caption{Application scenarios of vibrotactile captioning.}

\label{vibrotactile_caption}
\end{figure}

Despite their broad applicability, vibrotactile signals are inherently complex and noisy, which makes their interpretation challenging \cite{10452836}. To facilitate a deeper understanding of these signals, we adapt the concept of audio-visual captioning and introduce the task of \textit{vibrotactile captioning}, which translates these 1D signals into structured natural language. Rather than relying on low-level features or fixed taxonomies, natural language aligns more closely with human perception and facilitates intuitive understanding. As illustrated in Fig. \ref{vibrotactile_caption}, vibrotactile captioning enables three practical applications: 
(i)~\textbf{Semantic indexing}, where captions support language-based search, retrieval, and reasoning;
(ii)~\textbf{Material inspection}, where standardized textual summaries assist in quality control and verification within industrial workflows;
(iii)~\textbf{Virtual perception}, where captions offer semantic guidance to augment texture understanding in virtual reality environments with limited haptic resolution. Although Large Language Models (LLMs) have shown strengths in describing signals, their massive parameter and computational power requirements also limit their applications in the above scenarios.

To the best of our knowledge, vibrotactile captioning or similar works have not been explored. This might be attributed to data scarcity and technical challenges. \textbf{Data Scarcity.} First, there are few publicly available vibrotactile datasets, and most lack natural language annotations. For example, the LMT haptic texture database \cite{1474014} offers triaxial acceleration signals but does not provide textual descriptions, reducing its value for cross-modal learning. Second, vibrotactile signals often contain relevant and irrelevant components, making it difficult to extract clean and meaningful features \cite{s17112653}. Third, human descriptions of tactile sensations vary widely depending on individual perception, expertise, and attention \cite{PatioLakatos2019FromVS}, which hinders consistent large-scale annotation. \textbf{Technical Challenges.} First, captioning models developed for image, video, or audio domains rely on structural priors such as spatial layouts, motion continuity, or acoustic regularity. The absence of these priors in vibrotactile signals leads to difficulties in cross-modal transfer. Second, vibrotactile data encodes material surface properties as complex temporal dynamics that combine periodic patterns ({\it e.g.}, from regular textures) and aperiodic patterns ({\it e.g.}, from irregular surfaces or noise). This hybrid structure poses significant modeling challenges for single-stream encoders. 

To address the above issues, we propose ViPAC, a captioning framework tailored to vibrotactile signals. To mitigate data scarcity, we employ GPT-4o to generate textual descriptions of material surface images and pair them with their corresponding vibration signals to construct a cross-modal dataset. To overcome modeling challenges, we design a dual-branch encoder that separately processes periodic and aperiodic components for stable and transient vibrotactile cues, respectively. These features are fused via adaptive weighting and decoded via a Transformer-based decoder. This design allows ViPAC to effectively model complex vibrotactile structures and generate perception-aligned descriptions. Our main contributions are as follows:

\begin{itemize}
    \item[$\bullet$] We introduce the task of {\it vibrotactile captioning}, which aims to convert 1D triaxial acceleration signals into structured natural language descriptions that reflect characteristics of material surface and haptic signals. This task establishes a new direction for semantic modeling and interpretation of tactile data.

    \item[$\bullet$] We propose {\it ViPAC}, the first captioning framework specifically designed for vibrotactile signals. The model employs a dual-branch encoder to separately extract periodic and aperiodic features, and integrates them through an adaptive fusion mechanism that captures the hybrid temporal structure of tactile inputs. The fused representation is then decoded into natural language using a standard Transformer decoder.

    \item[$\bullet$] We construct {\it LMT108-CAP}, a vibrotactile-text paired dataset generated using GPT-4o under controlled linguistic constraints. Comprehensive experiments on the dataset validate the effectiveness of ViPAC in capturing the temporal structure of tactile signals and producing descriptive outputs that align with human interpretations.

\end{itemize}

The remainder of this paper is organized as follows. Section II reviews related work on vibrotactile datasets and text description tasks. Section III introduces the construction of the proposed LMT108-CAP vibrotactile–text paired dataset. Section IV presents the ViPAC framework, including the dual-branch encoder, dynamic fusion mechanism, and decoding strategy. Section V reports experimental results, ablation studies, and a retrieval demo to validate the effectiveness of the proposed method. Finally, Section VI concludes the paper and outlines directions for future work.

\section{Related Work}

\subsection{Datasets}

Existing vibrotactile datasets fall into two categories: visuotactile patterns and multiple 1D signals. Significant progress has been made in visuotactile datasets such as TVL~\cite{10.5555/3692070.3692632} and Touch100k~\cite{touch100k}, which use vision-based sensors to produce representations compatible with computer vision techniques, thereby facilitating multimodal alignment and captioning. In contrast, multiple 1D vibrotactile datasets aligns with the IEEE P1918.1 standard that advocates multiple 1D vibration signals as the canonical format. For example, the LMT haptic texture database~\cite{1474014} provides triaxial acceleration signals without paired natural language descriptions, which restricts its applicability in cross-modal learning. To date, no public dataset aligns triaxial signals with textual descriptions, posing significant obstacles to model training and evaluation. Motivated by the success of LLMs in visuotactile research \cite{10.5555/3692070.3692632}, we extend this paradigm to multiple 1D vibrotactile signals. We use GPT-4o to generate natural language descriptions for surface images and pair them with corresponding triaxial acceleration signals, resulting in a new dataset that supports tactile semantic modeling and cross-modal learning.

\begin{figure*}[t!]
    \centering
    \includegraphics[width=1\linewidth]{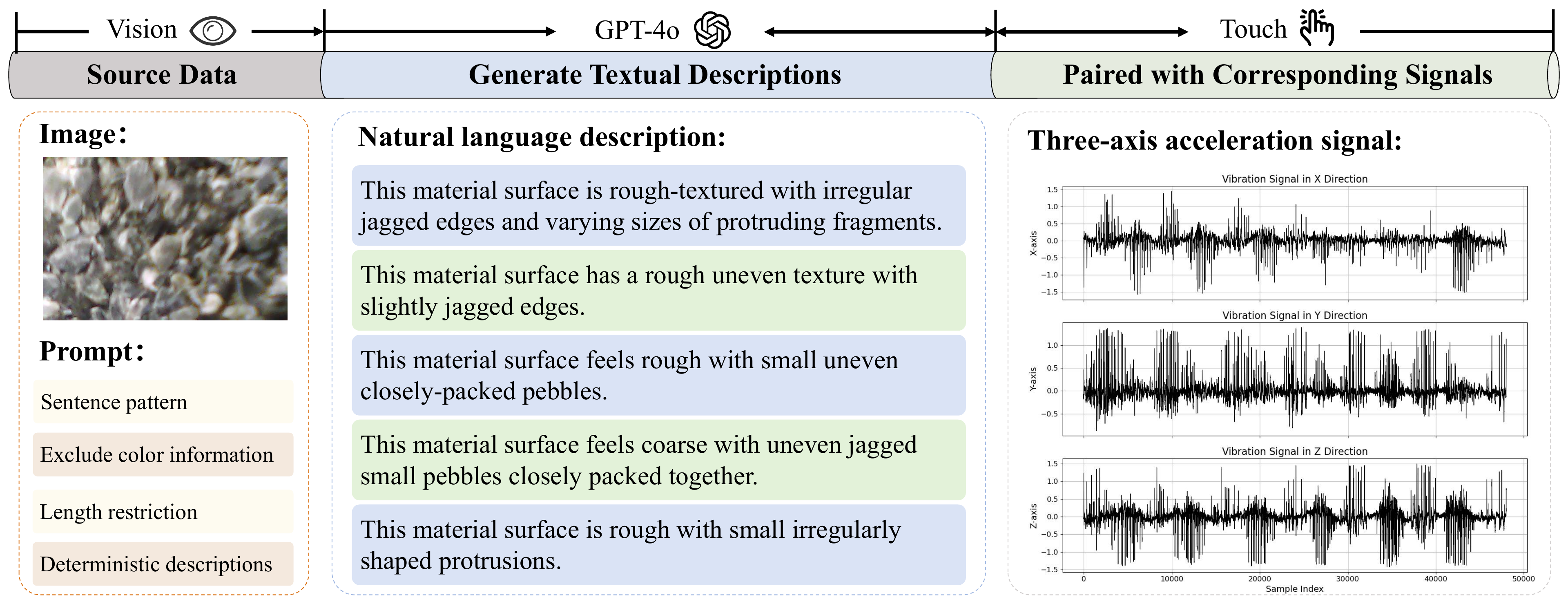}
    \caption{Illustration of the vibrotactile-text dataset generation process. Surface images from the LMT-108 dataset are provided as input to GPT-4o, which generates five textual descriptions per image under predefined linguistic constraints. These descriptions are then paired with the corresponding triaxial acceleration signals collected from the same material surfaces, resulting in the final vibrotactile-text dataset.}
    \label{gen_data}
\end{figure*}

\begin{figure}[t]
  \centering
  \includegraphics[width=\linewidth]{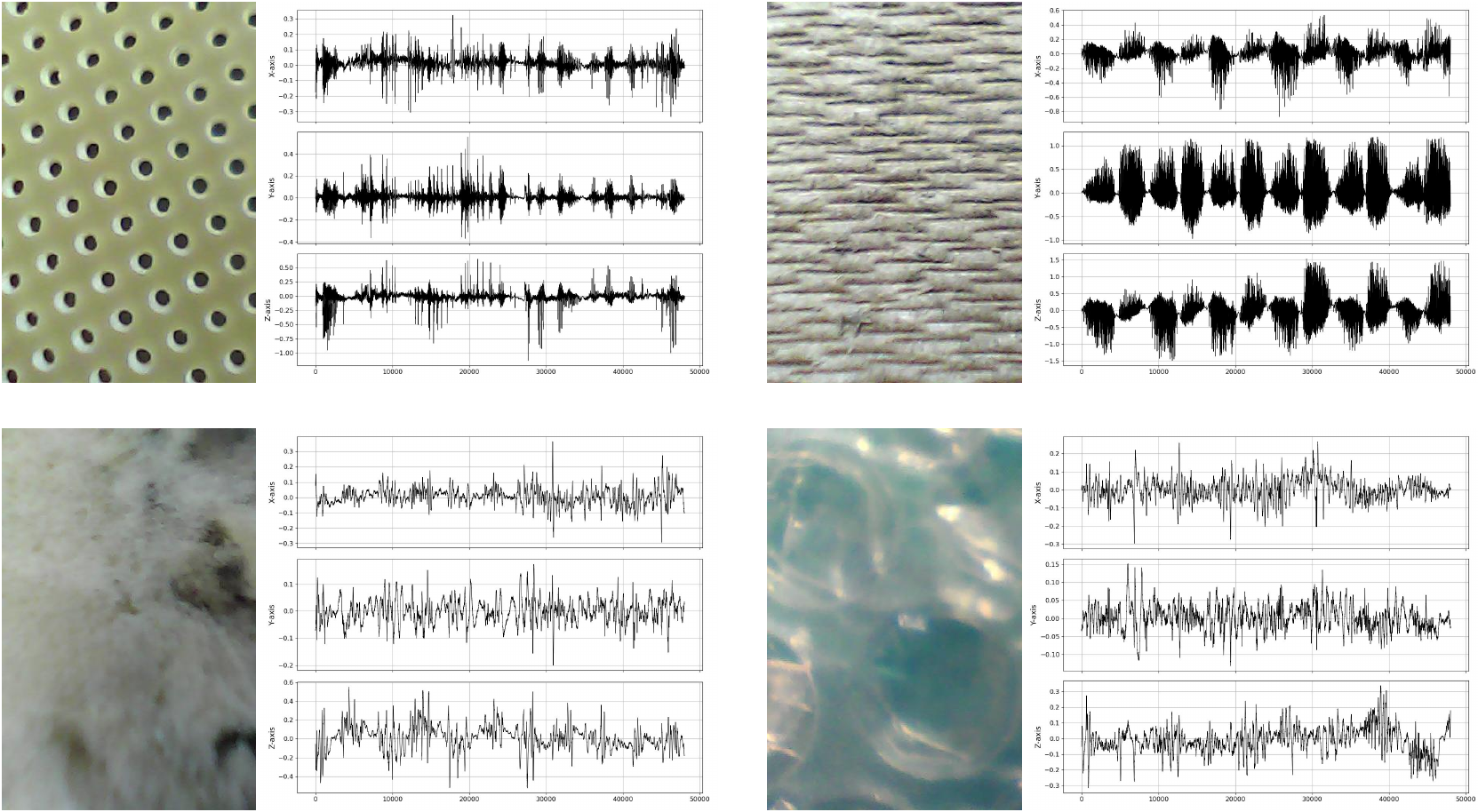}
  \caption{Examples of material surface images and their corresponding three-axis vibrotactile signals. Top: materials with regular textures exhibit strong periodicity. Bottom: irregular surfaces yield noisy, aperiodic signals. This motivates the use of distinct modeling pathways.}
  \label{dataset}
\end{figure}

\subsection{Text Description Tasks}

\subsubsection{Image-to-Text}

Image captioning aims to generate detailed textual descriptions of visual content by leveraging the structured spatial semantics present in images, such as object presence, spatial layout, and visual attributes~\cite{9410374, 9869686, Kornblith_2023_ICCV}. These methods perform well in visual domains but cannot be directly applied to vibrotactile data. Unlike images, vibrotactile signals do not contain spatial structure and instead encode material properties through temporal and spectral patterns. This fundamental difference makes conventional vision-based feature extractors, such as Convolutional Neural Networks (CNNs), ineffective when applied to multiple 1D tactile signals.

\subsubsection{Video-to-Text}

Video captioning extends image captioning into the temporal domain, modeling dynamic content such as motion trajectories, temporal coherence, and scene transitions~\cite{10.1007/978-3-030-58568-6_13, 9578758, 9578069}. Although vibrotactile signals also evolve over time, they capture localized, fine-grained surface interactions rather than global scene-level dynamics. As a result, video captioning models, which emphasize macroscopic visual motion, fail to represent the subtle frequency-based variations characteristic of tactile vibrations.

\subsubsection{Audio-to-Text}

Audio captioning generates textual descriptions from acoustic signals, benefiting from well-defined semantic categories such as speech, music, and environmental sounds, along with consistent recording conditions~\cite{10.1109/TASLP.2024.3416686, deshmukh2024training, 10572302}. In contrast, vibrotactile signals exhibit high variability due to scanning speed and contact force, and lack standardized semantic labels. This variability introduces ambiguity and noise, making direct application of audio captioning models ineffective for vibrotactile interpretation.

Above all, these modality-specific captioning models rely on structural priors that do not hold in the vibrotactile domain. Image models assume spatial semantics, video models emphasize motion continuity, and audio models exploit categorical regularity—none of which apply to tactile vibrations. To address this gap, we propose ViPAC, a captioning framework that models vibrotactile-specific features through separate periodic and aperiodic branches. The model further employs adaptive fusion to accommodate temporal variability and mitigate semantic ambiguity. This design provides a dedicated solution to the challenges of cross-modal captioning in tactile contexts.

\begin{figure*}[t!]
\centering
\includegraphics[width=1\textwidth]{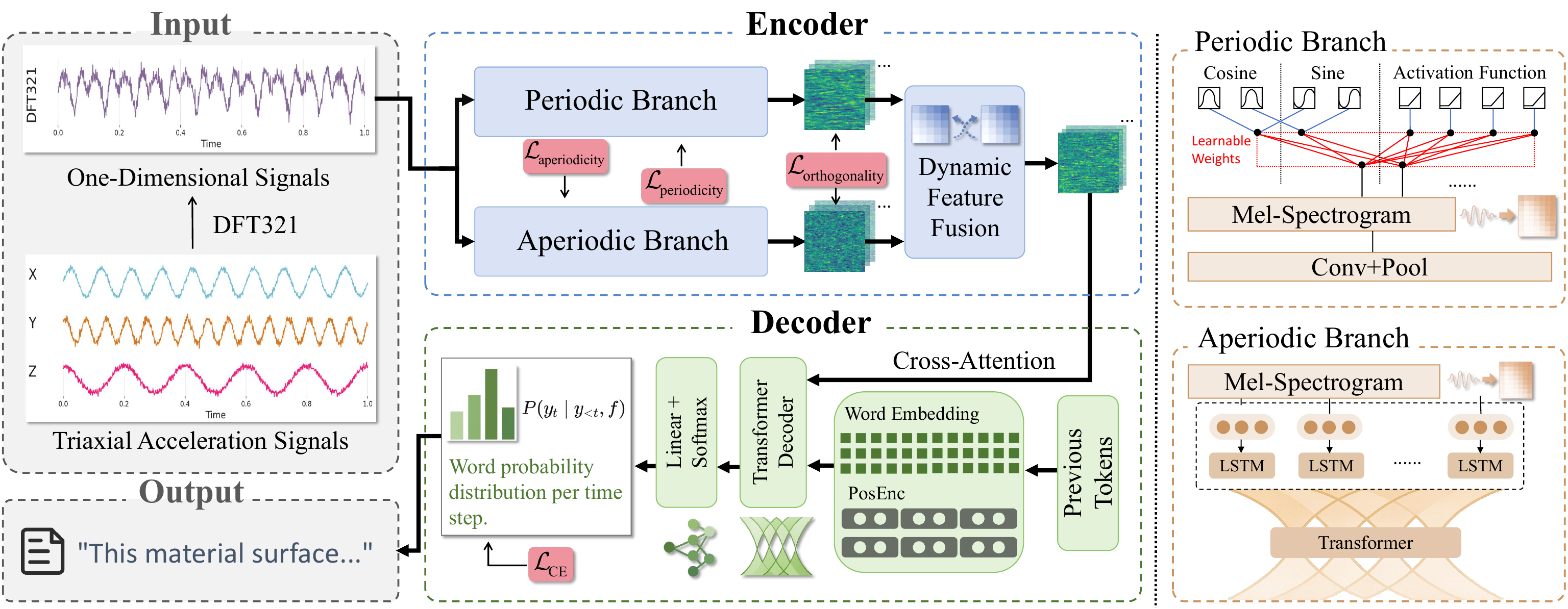} 
\caption{ViPAC takes triaxial acceleration signals as input and applies DFT321 to obtain 1D vibration data. These signals are processed by a dual-branch encoder that separately models periodic and aperiodic components using FAN-based frequency analysis and Transformer+LSTM-based temporal modeling, respectively. The extracted features are dynamically fused based on estimated periodicity scores, and the fused representation is decoded into natural language using a Transformer decoder.
}
\label{fig1}
\end{figure*} 

\section{LMT108-CAP Dataset Construction}

To overcome the lack of publicly available paired vibrotactile-text data, we introduce LMT108-CAP, a novel captioned vibrotactile dataset. This dataset, is based on the LMT-108 Surface-Materials database~\cite{7737070} promoted by the IEEE P1918.1 workgroup. The LMT-108 Surface-Materials database contains 108 distinct material surfaces grouped into 10 categories, each recorded with triaxial acceleration, friction, sound, and surface image data. Each surface is measured 20 times, resulting in 2,160 vibrotactile samples represented as 1D triaxial acceleration signals. The dataset is widely recognized and applied for its representativeness. As one of the field’s most widely used tactile benchmarks, LMT-108 offers representative coverage and thus provides a suitable basis for constructing paired data.

We construct the vibrotactile-text pairs with the following approach, which is inspired by image captioning dataset construction such as Flickr8K~\cite{10.5555/2566972.2566993}. Specifically, we employ GPT-4o to generate five textual descriptions for each material surface image in the LMT-108 dataset. The generation process adheres to four constraints designed to ensure relevance and consistency with the characteristics of vibrotactile signals:  
(i) \textbf{Sentence Pattern}: Descriptions start with ``This material surface...'' to encourage vibrotactile-focused textual output. (ii) \textbf{Exclude Color Information}: To align with vibrotactile signal characteristics, which do not capture color. (iii) \textbf{Length Restriction}: Each description contains no more than 15 words. (iv) \textbf{Deterministic Descriptions}: To prevent uncertain or imaginative associations, ensuring consistency and relevance to vibrotactile properties. Images are used only once to bootstrap the textual side; no visual inputs are used during training or inference, so the model learns signal–semantic correspondences rather than reproducing visual semantics.

LMT108-CAP contains surface images, their corresponding triaxial acceleration signals, and GPT-generated constrained captions, as illustrated in Fig. \ref{gen_data}. In total, the dataset includes 2,160 samples, each with five captions, yielding 10,800 paired instances following the five-caption protocol of Flickr8K \cite{10.5555/2566972.2566993}. We split the data into training and testing subsets using a 7:3 ratio, with 1,512 training samples (7,560 captions) and 648 test samples (3,240 captions), and perform corpus-level vocabulary control by filtering singleton words so that each remaining token appears at least once in both splits \cite{drossos2020clotho}. The selected materials span a broad range of textures, providing representative coverage for training and evaluating vibrotactile captioning models. To ensure the reliability of the generated text, we impose strict post-processing constraints. The LMT-108 dataset has also been widely used in prior work on tactile sensing and surface analysis \cite{9479777, 9018269, 10.1007/978-3-319-93399-3_3}, further supporting its suitability as a foundation for paired caption construction.

\section{Proposed ViPAC Model}

The primary innovation of this work lies in defining and systematically tackling the novel task of vibrotactile captioning. Vibrotactile signals, unlike visual or auditory signals, often combine periodic and aperiodic components that arise from structured (periodic) and unstructured (aperiodic) surface interactions \cite{PAWLUS2020106530}, as illustrated in Fig. \ref{dataset}. This hybrid structure challenges single-stream models and motivates an approach that treats these components explicitly.

We propose ViPAC, a dedicated framework that translates vibration signals into natural language through three stages: (i) dual-branch encoding to extract periodic and aperiodic features, (ii) adaptive fusion guided by signal characteristics, and (iii) sequence generation with a Transformer-based decoder. This design targets both repetitive and transient patterns in vibrotactile signals; the overall architecture is shown in Fig. \ref{fig1}.

In the encoder, periodic and aperiodic components are modeled separately to match their distinct statistics—an approach consistent with practices in surface engineering and tactile sensing. Concretely, the periodic branch employs a Fourier Analysis Network (FAN) for frequency-domain processing of stable, repeating patterns, while the aperiodic branch uses an LSTM+Transformer stack to capture irregular, long-range temporal variation. The two streams are then fused adaptively via a periodicity embedding \( p_i \), enabling the model to emphasize the appropriate branch for each input and to form a unified representation for decoding.

\subsection{Encoder Design}

To simplify the input while preserving perceptually relevant cues, we apply DFT321 to compress triaxial acceleration into a single 1-D signal. DFT321 is a widely used, standard front end in vibrotactile processing, including work that processes the LMT-108 dataset in the same way \cite{9479777}, and it is consistent with the IEEE P1918.1 recommendations. Landin et al. \cite{10.1007/978-3-642-14075-4_12} introduced DFT321 and showed that collapsing triaxial high-frequency vibration into a single magnitude channel causes no noticeable perceptual degradation. Our objective is to generate semantic descriptions of surface properties; the relevant cues are texture statistics and spectral structure rather than full 3-D force or direction vectors. Direction-invariant magnitude fusion therefore removes orientation nuisance while preserving the information needed for captioning.

The resulting signal $t_i$ typically consists of both periodic patterns from regular textures and aperiodic fluctuations caused by irregular surfaces or noise:
\begin{equation}
t_i = t_{\text{PER}} + t_{\text{APER}}.
\end{equation}

Conventional single-stream encoders fail to adequately capture this hybrid structure:
\begin{equation}
\mathbf{f} = f_{\text{enc}}(t_i).
\end{equation}

To address this, we propose a dual-branch encoder that models periodic and aperiodic components independently:
\begin{equation}
\mathbf{f}_{\text{PER}} = f_{\text{PER}}(t_i), \quad \mathbf{f}_{\text{APER}} = f_{\text{APER}}(t_i).
\end{equation}

These features are then fused into a unified representation:
\begin{equation}
\mathbf{f} = \text{Fuse}(\mathbf{f}_{\text{PER}}, \mathbf{f}_{\text{APER}}).
\end{equation}

\paragraph{Periodic Branch}

To capture stable, repetitive patterns that emerge from regular textures, we employ the Fourier Analysis Network (FAN)~\cite{dong2024fanfourieranalysisnetworks}. FAN is chosen for its effectiveness in extracting dominant frequency components from time-series signals. The transformed output is converted into a Mel-Spectrogram and further processed by a convolution-pooling module to obtain spectral representations:
\begin{equation}
    \mathbf{f}_{\text{PER},i} = \text{ConvPool}(\text{Mel}(\text{FAN}(t_i))).
\end{equation}
To enhance the learning of periodic structure, we introduce a periodicity loss computed as the variance of peak intervals in the autocorrelation function:
\begin{equation}
    \mathcal{L}_{\text{periodicity}} = \text{Var}(\Delta \text{FAN}(t_i)),
\end{equation}
where \(\Delta t\) denotes the lag between autocorrelation peaks.

\paragraph{Aperiodic Branch}

To model irregular, non-repetitive variations commonly observed in natural surfaces, we use a Transformer encoder combined with a Long-Short Term Memory (LSTM) layer. The layer captures short-term dynamics, while the Transformer provides long-range temporal modeling. This architecture is selected to address both local variability and global dependencies:
\begin{equation}
    \mathbf{f}_{\text{APER},i} = \text{Transformer}(\text{LSTM}(\text{Mel}(t_i))).
\end{equation}
To regularize the aperiodic features and avoid overly large activations, we apply a Mean Squared Error (MSE) style penalty to their magnitude:
\begin{equation}
\mathcal{L}_{\text{aperiodicity}} = \frac{1}{D} \left\| \mathbf{f}_{\text{APER},i} \right\|_2^2.
\end{equation}

\paragraph{Feature Decoupling and Fusion}

To ensure that the two branches capture complementary information, we apply an orthogonality loss between periodic and aperiodic features:
\begin{equation}
    \mathcal{L}_{\text{orthogonality}} = \| \langle \mathbf{f}_{\text{PER},i}, \mathbf{f}_{\text{APER},i} \rangle \|^2.
\end{equation}

We compute a scalar fusion weight $w_i \in [0,1]$ based on the periodicity embedding $\mathbf{p}_i$ via a sigmoid activation:
\begin{equation}
    w_i = \sigma(\alpha \cdot (\mathbf{p}_i - \tau)),
\end{equation}
where $\tau$ is a learnable threshold and $\alpha$ controls the sharpness of the transition. The final fused representation is:
\begin{equation}
    \mathbf{f}_i = w_i \cdot \mathbf{f}_{\text{PER},i} + (1 - w_i) \cdot \mathbf{f}_{\text{APER},i}.
\end{equation}

\subsection{Decoder Design}

The decoder transforms the fused tactile feature \( \mathbf{f}_i \) into a natural language description \( c_i = \{c_1, c_2, ..., c_T\} \) in an auto-regressive manner. We adopt a standard Transformer-based decoder~\cite{Mei2021act} consisting of a word embedding layer, a Transformer block with masked self-attention and cross-attention, and a linear projection layer with softmax activation.

At each time step \( t \), the decoder computes the probability of generating the next token conditioned on the previously generated tokens and the tactile feature:
\begin{equation}
p(c_t \mid c_{1:t-1}, \mathbf{f}_i, \theta) = \text{TransDec}(c_{1:t-1}, \mathbf{f}_i; \theta),
\end{equation}
where \( \theta \) denotes the decoder parameters. The sequence is trained to minimize the cross-entropy loss between the predicted and reference captions:

\begin{equation}
\mathcal{L}_{\text{CE}}(\theta) = -\frac{1}{T} \sum_{t=1}^{T} \log p(c_t \mid c_{1:t-1}, \mathbf{f}_i, \theta).
\end{equation}

During training, we apply the teacher forcing strategy, where the ground-truth tokens \( c_{1:t-1} \) are used as input to predict \( c_t \). While the decoder architecture itself is not novel, it plays a vital role in translating temporal tactile features into coherent, perception-aligned textual descriptions.

\subsection{Loss Functions}

The model is trained using a composite loss function that supervises both the tactile feature encoding and the text generation process. The overall objective is defined as:
\begin{align}
    \mathcal{L}_{\text{total}} &= \mathcal{L}_{\text{CE}} + \lambda_1 \mathcal{L}_{\text{periodicity}} + \lambda_2 \mathcal{L}_{\text{aperiodicity}} \notag \\
    &\quad + \lambda_3 \mathcal{L}_{\text{orthogonality}},
\end{align}
where \( \lambda_1, \lambda_2, \lambda_3 \) are hyperparameters controlling the relative importance of each term.

During test, ViPAC processes a raw triaxial acceleration signal by computing periodic and aperiodic features, fusing them adaptively, and decoding a textual description conditioned on the fused representation and previously generated tokens.

\section{Experiments}

\begin{table*}[htbp]
\setlength{\tabcolsep}{9pt}
\centering
\caption{Performance comparison of different captioning models.}
\renewcommand{\arraystretch}{1.3}
\begin{tabular}{l|cccccccccc}
\toprule
Model & BLEU1 & BLEU2 & BLEU3 & BLEU4 & ROUGE-L & METEOR & CIDEr & SPICE & SPIDEr \\
\midrule
ACT (DCASE 2021)     & 0.7204 & 0.5619 & 0.4498 & 0.3510 & 0.5931 & 0.2882 & 0.3107 & 0.2454 & 0.2780 \\
Kim et al. (ICASSP 2023)     & 0.7365 & 0.5732 & 0.4278 & 0.3086 & 0.5163 & 0.2597 & 0.7428 & 0.1942 & 0.4685 \\
Recap (ICASSP 2024)  & 0.7524 & 0.5891 & 0.4415 & 0.3213 & 0.5278 & 0.2714 & 0.7652 & 0.2126 & 0.4889 \\
ClipCap (2021) & 0.7343 & 0.5792 & 0.4667 & 0.3748 & 0.5819 & 0.2985 & 0.4737 & 0.2480 & 0.4959 \\
ViECap (ICCV 2023)  & 0.7432 & 0.5764 & 0.4573 & 0.3682 & 0.5867 & 0.2946 & 0.6981 & 0.2413 & 0.4697 \\
RCMF (TCSVT 2024) & 0.7480 & 0.5900 & 0.4520 & 0.3350 & 0.5800 & 0.2750 & 0.7200 & 0.2260 & 0.4730 \\
ViPAC  & \textbf{0.7635} & \textbf{0.6005} & \textbf{0.4782} & \textbf{0.3861} & \textbf{0.6047} & \textbf{0.3094} & \textbf{0.7795} & \textbf{0.2592} & \textbf{0.5194} \\
\bottomrule
\end{tabular}
\label{tab:results}
\end{table*}

\begin{figure*}[t!]
\centering
\includegraphics[width=1\textwidth]{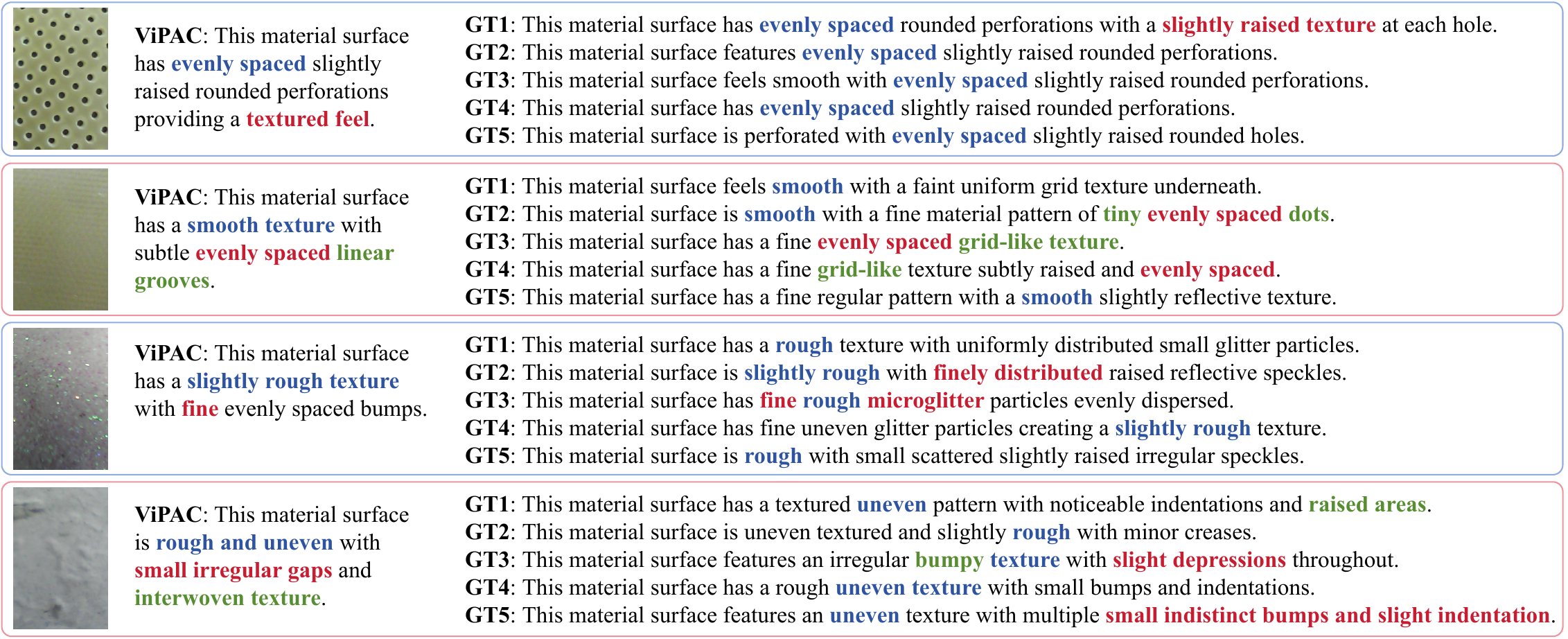} 

\caption{Qualitative comparisons between ViPAC generated captions and five GPT-4o reference descriptions for four representative materials. Matched phrases are highlighted to emphasize semantic consistency. The selected samples—covering regular perforations, fine grids, rough glitter, and irregular bumps—demonstrate ViPAC’s ability to produce accurate and diverse textual descriptions directly from vibrotactile signals.}

\label{result}
\end{figure*}

\begin{figure}[t!]
    \centering
    \includegraphics[width=1\linewidth]{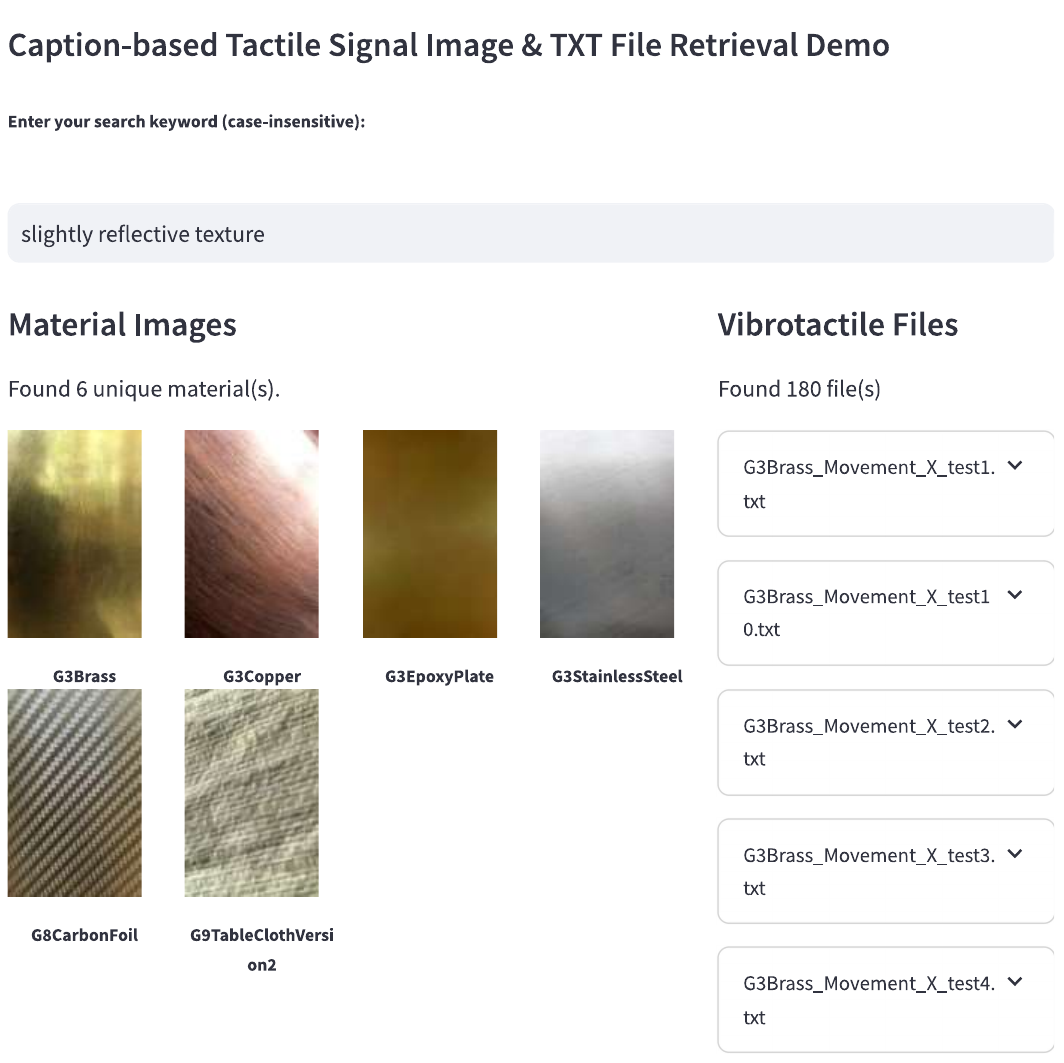}
    \caption{Demo interface for caption-based material retrieval. The left panel displays matched material images; the right panel shows vibrotactile files. Images are used only for visualization.}
    \label{fig:demo}
\end{figure}

\subsection{Experimental Setup}

{\bf Compared Methods.} To validate the effectiveness of our ViPAC framework, we compared it with existing methods from audio captioning tasks, including ACT \cite{Mei2021act}, Kim et al. \cite{kim2023prefix}, and Recap \cite{10448030}, as well as image captioning methods adapted for vibrotactile signals via Mel-spectrogram conversion, such as ClipCap \cite{mokady2021clipcap}, ViECap \cite{Fei_2023_ICCV} and RCMF \cite{10589672}. This cross-domain comparison rigorously evaluates our framework’s ability to translate vibrotactile vibration signals into natural language under diverse methodological paradigms. All methods were trained under the same experimental conditions, including the same training, validation, and test data splits. Although ClipCap \cite{mokady2021clipcap} was not published through a peer-reviewed venue, it has gained popularity due to its novel approach of injecting CLIP-based visual embeddings as prefix tokens into pretrained language models for image captioning.


{\bf Evaluation Metrics.} We adopted five widely used metrics to comprehensively evaluate caption quality. BLEU \cite{papineni-etal-2002-bleu} measures $n$-gram overlap using the geometric mean of modified precision with a brevity penalty. ROUGE-L \cite{lin-och-2004-automatic} computes an F-score based on the longest common subsequence, reflecting structural alignment. METEOR \cite{denkowski-lavie-2014-meteor} improves recall-oriented evaluation by incorporating synonym and stem matching. CIDEr \cite{7299087} captures semantic relevance via TF-IDF-weighted cosine similarity of $n$-grams. SPICE \cite{10.1007/978-3-319-46454-1_24} evaluates semantic content by comparing scene graph tuples (objects, attributes, relations). SPIDEr, the arithmetic mean of SPICE and CIDEr, balances semantic accuracy and lexical fidelity.

{\bf Experimental Configuration.} All experiments were conducted on a Windows 10 64-bit workstation equipped with an Intel Core i7-14700KF CPU (3.40 GHz), 32 GB RAM, and an NVIDIA GeForce RTX 4090 D GPU (24 GB). Model development and training were implemented using Python 3.7.12, PyTorch 2.1.0, and CUDA Toolkit 11.1.


\begin{table*}[htbp]
\setlength{\tabcolsep}{7pt}
\centering
\caption{Ablation study on model components. We evaluate the effects of removing the periodic branch, aperiodic branch, and adaptive fusion module. The full model consistently achieves the highest performance, confirming that both branches and the dynamic weighting mechanism contribute to the effectiveness of ViPAC.}
\renewcommand{\arraystretch}{1.6}
\setlength\tabcolsep{9pt}
\begin{tabular}{l|ccccccccc}
\midrule
Variant & BLEU1 & BLEU2 & BLEU3 & BLEU4 & ROUGE-L & METEOR & CIDEr & SPICE & SPIDEr \\
\midrule
Periodic Only         & 0.6124 & 0.4893 & 0.3582 & 0.2645 & 0.5012  & 0.2287 & 0.4215 & 0.1683 & 0.2949 \\
Aperiodic Only     & 0.6657 & 0.5328 & 0.4127 & 0.3061 & 0.5237  & 0.2614 & 0.5362 & 0.1947 & 0.3655 \\
No Fusion & 0.6829 & 0.5535 & 0.4312 & 0.3198 & 0.5379  & 0.2745 & 0.5683 & 0.2054 & 0.3869 \\
\textbf{ViPAC (Full)}    & \textbf{0.7635} & \textbf{0.6005} & \textbf{0.4782} & \textbf{0.3861} & \textbf{0.6047} & \textbf{0.3094} & \textbf{0.7795} & \textbf{0.2592} & \textbf{0.5194} \\
\midrule
\end{tabular}

\label{tab:ablation_study}
\end{table*}

\begin{table*}[t]
\centering
\caption{Complete evaluation metrics when a specific material category (G1 to G9) is excluded from training. Each row corresponds to a model trained without the respective category.}
\renewcommand{\arraystretch}{1.4}
\setlength\tabcolsep{9pt}
\begin{tabular}{c|ccccccccc}
\toprule
\textbf{Excluded Category} & \textbf{BLEU1} & \textbf{BLEU2} & \textbf{BLEU3} & \textbf{BLEU4} & \textbf{ROUGE-L} & \textbf{METEOR} & \textbf{CIDEr} & \textbf{SPICE} & \textbf{SPIDEr} \\
\midrule
G1 & 0.6325 & 0.4674 & 0.3592 & 0.2841 & 0.5035 & 0.2394 & 0.4892 & 0.1553 & 0.3223 \\
G2 & 0.7152 & 0.5483 & 0.4301 & 0.3456 & 0.5658 & 0.2765 & 0.6518 & 0.2169 & 0.4344 \\
G3 & 0.6950 & 0.5298 & 0.4164 & 0.3340 & 0.5536 & 0.2661 & 0.6142 & 0.2007 & 0.4075 \\
G4 & 0.6159 & 0.4563 & 0.3508 & 0.2756 & 0.4927 & 0.2328 & 0.4671 & 0.1475 & 0.3073 \\
G5 & 0.7524 & 0.5891 & 0.4667 & 0.3748 & 0.5931 & 0.2985 & 0.7100 & 0.2480 & 0.4380 \\
G6 & 0.6437 & 0.4779 & 0.3660 & 0.2901 & 0.5124 & 0.2450 & 0.5094 & 0.1625 & 0.3360 \\
G7 & 0.7053 & 0.5385 & 0.4235 & 0.3404 & 0.5597 & 0.2713 & 0.6332 & 0.2074 & 0.4203 \\
G8 & 0.5998 & 0.4427 & 0.3384 & 0.2667 & 0.4793 & 0.2260 & 0.4425 & 0.1394 & 0.2909 \\
G9 & 0.5412 & 0.3906 & 0.2975 & 0.2328 & 0.4351 & 0.1925 & 0.3764 & 0.1152 & 0.2458 \\
\textbf{Full Data} & \textbf{0.7635} & \textbf{0.6005} & \textbf{0.4782} & \textbf{0.3861} & \textbf{0.6047} & \textbf{0.3094} & \textbf{0.7795} & \textbf{0.2592} & \textbf{0.5194} \\
\bottomrule
\end{tabular}
\label{tab:missing}
\end{table*}

\begin{table*}[t]
\centering
\caption{Ablation on input representation with the full metric suite. Training three independent models with only one axis (X/Y/Z) yields consistently worse captions than using DFT321 to fuse triaxial signals into one channel.}
\renewcommand{\arraystretch}{1.35}
\setlength\tabcolsep{9pt}
\begin{tabular}{l|ccccccccc}
\midrule
\textbf{Input} & \textbf{B1} & \textbf{B2} & \textbf{B3} & \textbf{B4} & \textbf{R-L} & \textbf{MET} & \textbf{CIDEr} & \textbf{SPICE} & \textbf{SPIDEr} \\
\midrule
X-only        & 0.6820 & 0.5220 & 0.4010 & 0.3087 & 0.5450 & 0.2668 & 0.5954 & 0.2130 & 0.4042 \\
Y-only        & 0.7090 & 0.5450 & 0.4250 & 0.3370 & 0.5660 & 0.2790 & 0.6512 & 0.2068 & 0.4290 \\
Z-only        & 0.6640 & 0.5060 & 0.3890 & 0.2924 & 0.5380 & 0.2583 & 0.5627 & 0.2045 & 0.3836 \\
Mean(X,Y,Z)   & 0.6850 & 0.5240 & 0.4050 & 0.3127 & 0.5500 & 0.2680 & 0.6031 & 0.2081 & 0.4056 \\
\textbf{DFT321} & \textbf{0.7635} & \textbf{0.6005} & \textbf{0.4782} & \textbf{0.3861} & \textbf{0.6047} & \textbf{0.3094} & \textbf{0.7795} & \textbf{0.2592} & \textbf{0.5194} \\
\midrule
\end{tabular}
\label{tab:dft321_ablation_full}
\end{table*}

\subsection{Experimental Results}

{\bf Qualitative and Quantitative Analysis.} We evaluate ViPAC using standard captioning metrics and compare it against baseline methods adapted from image and audio domains. As shown in Table~\ref{tab:results}, ViPAC achieves the highest scores across all metrics, with particularly notable gains in CIDEr and SPICE, indicating improved lexical precision and semantic consistency. Fig.~\ref{result} illustrates qualitative comparisons between generated captions and GPT-4o-generated references for four representative material types: regular grid, fine mesh, glittery roughness, and bumpy irregularities. ViPAC reliably captures salient tactile features such as smoothness, periodic spacing, and surface irregularity. However, challenging textures occasionally result in lower semantic alignment, highlighting the need for improved sensitivity to subtle signal variations. These results collectively demonstrate the effectiveness of ViPAC in producing accurate, fluent, and perception-consistent textual descriptions of vibrotactile signals.

{\bf Demo of ViPAC Application in Retrieval.} To demonstrate a practical application of vibrotactile captioning, we designed a web-based demo interface that supports text-based retrieval over material samples. Users can input a keyword or sentence to retrieve all generated captions containing the query. The interface displays the corresponding material name and reference image in the left panel and the matched vibrotactile files in the right panel. This proof-of-concept system operates solely on vibrotactile input; no visual data are used during training or inference. Reference images are included only for visualization, helping users interpret the results. As shown in Fig.~\ref{fig:demo}, the interface supports both keyword and sentence queries and exemplifies the semantic indexing and retrieval scenario introduced earlier.

\subsection{Ablation Study}

{\bf Component Ablation.} We conduct ablation experiments to assess the impact of each model component. As shown in Table~\ref{tab:ablation_study}, the full ViPAC model achieves the highest performance across all metrics. Removing the dynamic fusion module and using fixed weights leads to noticeable degradation, confirming the effectiveness of adaptive weighting. The aperiodic-only variant performs better than the periodic-only one, indicating that irregular features contain more standalone information. These results validate that periodic and aperiodic features are complementary and that dynamic fusion is essential for modeling hybrid tactile structures.

{\bf Generalization under Missing Categories.} To assess generalization, we retrain nine models, each with one material category (G1–G9) held out, and evaluate on that category. Table~\ref{tab:missing} reports the full metric suite. Across all held-out categories, ViPAC remains stable. The strongest zero-shot transfer appears on G5, followed by G2 and G7; the weakest is G9, suggesting its textures are less represented by the remaining categories. Trends are consistent across lexical and semantic metrics, indicating that periodic/aperiodic cues learned from other groups transfer broadly. The “Full Data” row provides the upper bound when no category is removed.

{\bf Justifying DFT321 Axis Fusion.} To verify that compressing triaxial acceleration into a single magnitude channel via DFT321 is reasonable for captioning, we replace DFT321 with single-axis inputs and train three independent models using only $x$, only $y$, or only $z$ signals. All settings are identical to the main model. Table \ref{tab:dft321_ablation_full} reports the full metric suite. DFT321 clearly outperforms any single-axis alternative across lexical (BLEU/ROUGE/METEOR) and semantic metrics (CIDEr/SPICE/SPIDEr). The best single-axis model (Y-only) still trails DFT321 by 4.9 BLEU4 points and 0.090 SPIDEr; the mean of single-axis runs remains notably lower.

\section{Conclusion}
This work introduces vibrotactile captioning, which aims to translate multiple 1D triaxial acceleration signals into natural-language descriptions that convey material surface characteristics. To enable this task, we construct a paired vibrotactile-text dataset using an LLM and propose a dual-branch architecture that separately models the periodic and aperiodic components of vibrotactile signals. The proposed method, ViPAC, generates accurate and semantically rich captions, as demonstrated by both quantitative metrics and qualitative comparisons. We further develop a retrieval demo to showcase the usefulness of vibrotactile captioning for caption-based material search.
Limitations of this study include the modest dataset size, the lack of large-scale human-authored captions, and the reliance on triaxial acceleration signals only. Future work will expand the dataset with more diverse materials and human-verified descriptions, improve robustness under real-world conditions, and explore multimodal fusion with additional sensory signals.

\bibliographystyle{IEEEtran}
\bibliography{tmmbib}





\vfill

\end{document}